\documentclass[journal]{new-aiaa}
\usepackage[utf8]{inputenc}
\usepackage{textcomp}

\usepackage{graphicx}
\usepackage{amsmath}
\usepackage[version=4]{mhchem}
\usepackage{siunitx}
\usepackage{longtable,tabularx}
\usepackage{soul}
\usepackage{color}
\title{Multi-Class Multiple Instance Learning for Predicting Precursors to Aviation Safety Events}

\author{Marc-Henri Bleu Laine\footnote{Graduate Research Assistant, Aerospace Systems Design Laboratory, School of Aerospace Engineering, AIAA Student Member}, Tejas G. Puranik\footnote{Research Engineer II, Aerospace Systems Design Laboratory, School of Aerospace Engineering, AIAA Senior Member}, Dimitri N. Mavris\footnote{S.P. Langley NIA Distinguished Regents Professor and Director of Aerospace Systems Design Laboratory, School of Aerospace Engineering, AIAA Fellow}}\affil{Georgia Institute of Technology, Atlanta, GA, 30332}
\author{Bryan Matthews\footnote{Data Science Research Engineer}}
\affil{KBR Inc., NASA Ames Research Center}

\begin{document}

\maketitle

\begin{abstract}

In recent years, there has been a rapid growth in the application of machine learning techniques that leverage aviation data collected from commercial airline operations to improve safety. Anomaly detection and predictive maintenance have been the main targets for machine learning applications. However, this paper focuses on the identification of precursors, which is a relatively newer application. Precursors are events correlated with adverse events that happen prior to the adverse event itself. Therefore, precursor mining provides many benefits including understanding the reasons behind a safety incident and the ability to identify signatures, which can be tracked throughout a flight to alert the operators of the potential for an adverse event in the future. This work proposes using the multiple-instance learning (MIL) framework, a weakly supervised learning task, combined with carefully designed binary classifier leveraging a Multi-Head Convolutional Neural Network-Recurrent Neural Network (MHCNN-RNN) architecture. Multi-class classifiers are then created and compared, enabling the prediction of different adverse events for any given flight by combining binary classifiers, and by modifying the MHCNN-RNN to handle multiple outputs. Results obtained showed that the multiple binary classifiers perform better and are able to accurately forecast high speed and high path angle events during the approach phase. Multiple binary classifiers are also capable of determining the aircraft's parameters that are correlated to these events. The identified parameters can be considered precursors to the events and may be studied/tracked further to prevent these events in the future.
\end{abstract}

\clearpage
\section*{Nomenclature}
{\renewcommand\arraystretch{1.0}
\noindent\begin{longtable*}{@{}l @{\quad=\quad} l@{}}
$CNN$ & Convolutional Neural Network\\
$DT-MIL$ & Deep Temporal Multiple-Instance Learning\\
$FDR$ & Flight Data Recorder\\
$FN$ & False Negative\\
$FP$ & False Positive\\
$FOQA$ & Flight Operations Quality Assurance\\
$GRU$ & Gated Recurrent Unit\\
$MHCNN$ & Multi-Head Convolutional Neural Network\\
$MIL$ & Multi-Instance Learning\\
$QAR$ & Quick Access Recorder\\
$RNN$ & Recurrent Neural Network\\
% $ASOS$ & Automated Surface Observing Systems \\
% $ASPM$ & Aviation System Performance Metrics\\
% $ATC$ & Air Traffic Controllers\\
% $CASSIE$& Computing Analytics and Shared Services Integrated Environment \\
% $CSV$ & Comma-Separated Value\\
% $EDCT$ & Expected Departure Clearance Times \\
% $FIXM$  & Flight Information Exchange Model\\

% $GDP$  & Ground Delay Program \\
% $GS$  & Ground Stop \\
% $LGA$ & LaGuardia Airport\\
% $NAS$ & National Airspace System \\
% $SMOTE$ & Synthetic Minority Over-sampling Technique \\
% $TFMS$ & Traffic Flow Management System \\
% $TMI$ & Traffic Management Initiative \\
$TN$ & True Negative\\
$TP$ & True Positive\\
\end{longtable*}}

\section{Introduction}

\begin{figure}[hbt!]
 \centering % Always center your figures and tables.
 \includegraphics[width=0.7\linewidth]{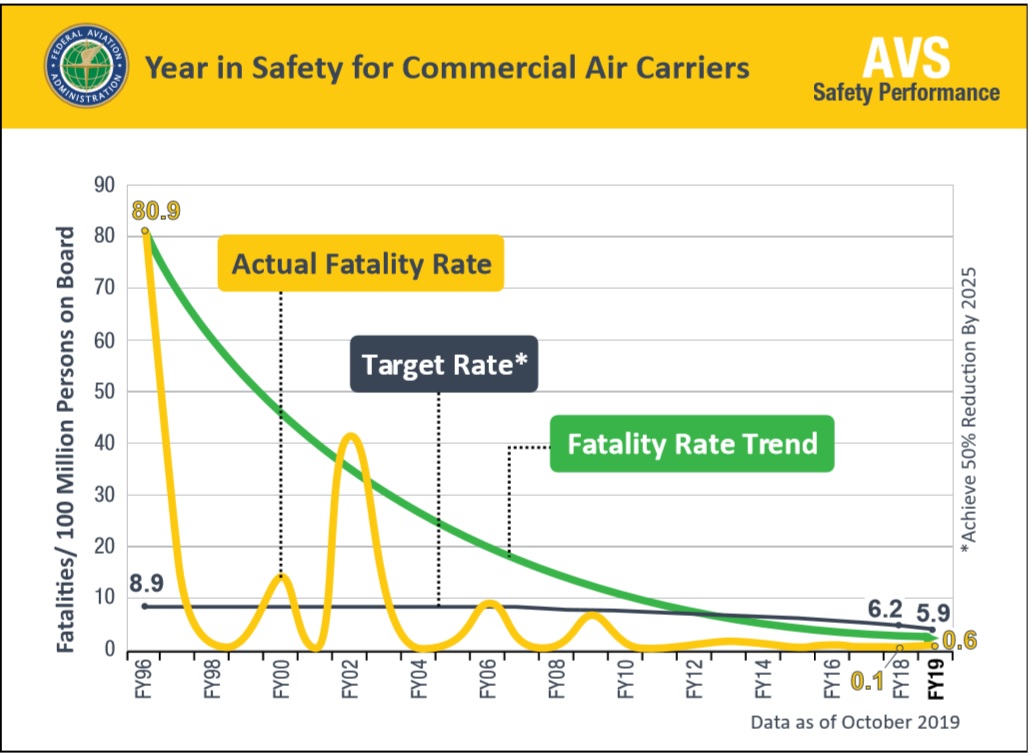}
 \caption{Commercial Aviation Fatalities from FY96 to FY19 \cite{FAA_safety}}
 \label{fig:fatalities}
\end{figure}

The aviation industry brings tremendous amount of social and economic benefits. It has been observed that the size of the air transportation industry has doubled every 15 years \cite{aviationbenefit2019}, and was expected to continue growing \cite{market_outlook} prior to the COVID-19 pandemic. In 2018 alone, airlines around the world carried a total of 4.3 billion passengers while the total global economic impact of the industry reached USD 2.7 Trillion in 2016 \cite{aviationbenefit2019}. Furthermore, even though the industry continues to grow and more people are flying everyday, aviation safety has improved over the past decades as seen on Figure~\ref{fig:fatalities}. 

The number of fatalities for commercial air carriers decreased from 81 per 100 million persons on board in 1996 to 0.6 in 2019, which is below the target rate set by the Federal Aviation Administration (FAA). This reduction is the result of the efforts undertaken by agencies such as the  National Aeronautics and Space Administration (NASA), the FAA, the National Transportation Safety Board (NTSB), and others. In particular, these efforts led to better certification standards, better operating procedures, and decision-support systems \cite{model_env_safety}. However, it is important to keep reducing the accident rate even further so that we do not observe a rise in the number of accidents given the industry's expected continued growth \cite{reduce_accident_rate_faa, market_outlook}. In order to significantly improve safety, the aviation industry has been moving towards a proactive approach to safety assessment and vulnerability identification, which consists of characterizing potential risks in terms of anomalies or deviations from nominal operations and the precursors to adverse events. This knowledge can be leveraged to increase awareness of emerging vulnerabilities amongst the operators and be incorporated into automated monitoring tools to flag the risks before they result in near-misses, incidents, or accidents \cite{aviation_safety_blog}. 

Extensive data collection and advancement in data-mining methodologies are key enablers to proactive risk management in aviation \cite{Logan2008, aviation_safety_blog}. Airline programs such as the Flight Operations Quality Assurance (FOQA) have enabled the creation of large, and heterogeneous data sets. FOQA is a voluntary safety program that is designed to make commercial aviation safer. Data is collected using devices such as Quick Access Recorder (QAR) or directly from Flight Data Recorder (FDR) \cite{FAA_FOQA}. Traditional techniques of flight data analysis have focused on a continuous cycle of data collection from on-board recorders, retrospective analysis of flight-data records, identification of operational safety exceedances, design and implementation of corrective measures, and monitoring to assess their effectiveness. Airlines work with the FAA to reduce and eliminate safety risks, and flight safety divisions within airlines generally use FOQA data to perform exceedence and statistical analyses. Exceedence analysis consist of setting specific limits to the recorded airborne data so that particular parameters that fall outside of the normal operating conditions can be flagged \cite{FAA_FOQA}. The level of exceedence can be programmed for different severities of events. These profiles are used to create distributions of various criteria, which enables airlines to evaluate flight risk levels and trend known vulnerabilities over time \cite{FAA_FOQA}. A validation step at the end of each analysis is performed to determine the nature of corrective actions required and to store valid events in databases for analyzing trends \cite{FAA_FOQA}. 

Research efforts have been made towards advancing the development of data-driven methodologies, applying data science techniques, and using modern machine learning, including deep learning algorithms, in the context of aviation safety \cite{anomaly_detection_review_survey, survey_data_mining, Matthews2013, Puranik2018_AD, Puranik2019_AD, Tong2018, deshmukh2019incremental}.
Majority of the data mining effort in aviation is directed at detecting anomalies in aviation data~\cite{MKL, flight_anomaly_dbscan_GT, anomaly_detection_gaussian, flight_anomaly_detection_real_time, Puranik2019_AD, Puranik2018_AD, survey_data_mining, vae, corrado} and leveraging unsupervised learning techniques due to lack of labeled data. Data mining has also been applied towards predictive maintenance in aviation~\cite{anomaly_detection_review_survey} and predicting future trajectory states~\cite{puranik2020_towards, Lee2021}. While identifying anomalies is important, it is also critical to investigate the causal factors or precursors to these anomalies or other safety events in order to understand them better and  prevent them in the future. Recent literature has focused on identifying precursors to safety events and anomalies using FOQA data~\cite{precursor_vijay, precursor_jamey, data_driven_precursor}. Precursors can be defined as any event that are correlated to a safety incident and occurs prior the incident itself \cite{precursor_vijay}. They are useful for forecasting safety events, and provide insights to why the event happened. Knowing the precursors can then be used to initiate actions to avoid events from occurring \cite{data_driven_precursor}. Therefore, being able to identify and monitor precursors is an important step towards proactive safety enhancement in aviation operations. One of the main challenges in identification of precursors observed in literature is the lack of subject-matter-expert validated labels for the identified anomalies and safety events. To account for the lack of labels usually encountered, this paper proposes the application of a weakly-supervised learning technique called multi-class multiple-instance learning (MIL)~\cite{li2018multiclass} to detect different multiple adverse events and their precursors.

Considering the above observations, the main aim of this paper is the development of a methodology that leverages highly dimensional aviation data to predict multiple adverse events and discover their precursors. This work will use a flight-level label called a bag-label in MIL terminology to predict the anomalies at the flight-level and identify their precursors along with their occurrence in-time during a flight. The current framework uses a deep learning model constructed as multi-headed convolutional neural network (CNN) where each flight sensor gets its own CNN and the multi-headed architecture is designed to work with the MIL framework. Successful implementation of the methodology uses the flight's data to 1) predict adverse events, and 2) determine the precursors to the predicted adverse events. One major benefit from this framework over current approaches will be the extension of binary classifier to perform multi-class predictions and the ability to retrieve the precursors with little to no post-processing. Additional benefits will include transferability of the model due its multi-head architecture, knowledge discovery to help analysts find root causes of safety events faster, and relative simplicity of the model to provide better transparency and explainability (compared to previous architectures).

\section{Background}
This section presents recent work conducted related to anomaly detection, precursor mining, and other applications of deep learning models in aviation safety. The data source utilized for this work is also presented along with the definition of multiple-instance learning and its accompanying assumptions.

\subsection{Previous Work}

Focusing on precursors to anomalies is important as it allows identifying potential causes to safety hazards. Recent work shows a growing interest in detecting precursors in various domains. Multiple techniques have been explored and the most relevant ones are summarized in this subsection.

Yue Ning et al.~\cite{precursor_MIL_text} presented an approach for precursor identification using nested Multi-Instance Learning (n-MIL). In their work, they forecast societal events in different cities. At the instance level, the probability of an  article published on a given day is modeled using a simple logistic function. These probabilities are then aggregated over a day, and finally the probabilities for different days are aggregated together up to a certain number of days before the event, creating the nested structure. The authors explain that the probability of a news article on a given day can be used to estimate how related an article is to the target. Since the articles with higher probabilities influenced the classifier decision, they are likely to be precursors of the predicted societal event. 

Janakiraman et al.~\cite{precursor_vijay} proposed to use a Deep Temporal Multiple-Instance Learning (DT-MIL) framework, which combines Multiple-Instance Learning and recurrent neural networks, in this case a gated recurrent unit (GRU), to mine precursors in FOQA data. In this approach, the individual time steps are considered low-level instances and the whole flight is considered a bag. The labels (occurrence of adverse event) are given at the bag level but the methodology takes advantage of MIL and uses the low-level instances to correctly predict the bag label, which allows to infer the instance-level label. The time-steps at which the probability of a safety event occurring is greater than a defined threshold are retrieved. The region of time for which the probability (called the precursor score) of a safety event occurring is high is then analyzed during post processing, where each feature is perturbed one at a time. The precursors are identified by finding the features whose perturbations had more significant impacts in reducing the precursor score. The DT-MIL model is also referred to as a newer version of the Automatic Discovery of Precursors in Time Series
(ADOPT) \footnote{\url{https://github.com/nasa/ADOPT}}, which is different from the architecture presented in \cite{precursor_vijay2}.

Ackley et al.~\cite{precursor_jamey} have used a sequential backward selection technique along with Random Forest classification models for predicting Unstable Approach adverse events. They have identified the critical parameters using a cumulative feature importance score and grouped them into various categories of parameters (such as energy-related, configuration-related, etc.) that contribute significantly towards the identification of the adverse event. Their analysis is conducted at fixed altitudes above the event detection trigger altitude of 1000 feet above touchdown. Similarly, Lee et al.~\cite{Lee2020} have also used a Random Forest algorithm to identify precursors to two different aviation safety events using supervised learning. The normalized precursor score is obtained using a Gini importance for all parameters contained in the classification model.

Melynk et al.~\cite{Melnyk2013_precursor} have proposed a framework for detecting precursors to aviation safety incidents due to human factors based on Hidden Semi-Markov Models. They performed an empirical evaluation of their models against traditional anomaly detection algorithms and demonstrated better performance on synthetic and flight simulator data. Mangortey et al.~\cite{Mangortey2020} used a variety of clustering techniques to identify clusters of nominal operations and subsequently determine important parameters that differentiate outliers from those nominal clusters. Despite presenting interesting insights, their method is still unsupervised and lacks validation. 

\subsection{Dashlink Flight Data}
The proposed framework is demonstrated using a publicly available data set obtained from NASA's DASHlink website, which is a collaborative sharing network for researchers in the Data Mining and Systems Health Management field\footnote{\url{https://c3.nasa.gov/dashlink/resources/?page=3&sort=-created&type=28}}. Flight data were recorded from a single type of regional jet operating in commercial service over a three-year period. The data contains detailed aircraft dynamics, system performance, and other engineering parameters but are de-identified such that it cannot be traced back to a particular manufacturer or airline. Since this data set is not part of any airline's FOQA program, additional preprocessing is required to create FOQA-like flags and label safety events for individual flights, where the labeling was created by using domain-based rules. The definitions of resulting adverse events are presented in Table~\ref{tab:adverse event}.

\begin{table}[hbt!]
\caption{Adverse Events Labeling}
\label{tab:adverse event}
\begin{center}
\begin{tabular}{|c|c|} \hline
 \textbf{Adverse Event} & \textbf{Comments} \\
\hline \hline
High Speed in Approach & Flagged at 1,000 ft \\ \hline
Low Speed in Approach & Flagged at 1,000 ft  \\ \hline
High Rate of Descent in Approach & Flagged between 1,000 - 500 ft  \\ \hline
High Bank in Approach &  Flagged between 1,000 - 400 ft  \\ \hline
High Path Angle in Approach & Flagged at 1,000 ft \\ \hline
Low Path Angle in Approach & Flagged at 1,000 ft \\ \hline
Deviation from Localizer & Flagged between 1,000 - 500ft \\ \hline
Deviation below Glideslope & Flagged between 1000 - 500ft \\ \hline
Flaps Late Setting at Landing  & None \\ \hline
\end{tabular}
\end{center}
\end{table}

Each of these events are characterized by a severity level ranging from 1 to 3, with 3 being the most severe. For this work, only flights without any safety events and the ones with safety events with a severity level 3 will be considered. This will allow limited overlaps between normal and abnormal operations. Moreover, to ensure a multi-class problem instead of a multi-label one, flights with multiple adverse events were omitted. The high speed in approach, and the high path angle in approach events were selected for this work because of their higher frequency in the data set. It is noted that the methodology developed in this work is applicable to any similar data set containing time series data of aircraft parameters and the known occurrence of an event. 

\subsection{Multiple-Instance Learning}

The multiple-instance learning framework is used when labels are only available in sets called bags \cite{MIL}, and each bag contains many instances. The learning task is therefore supervised, but since the labels are not provided for each instance, we say that the task is weakly supervised. This framework is of interest because it alleviates the burden of weak supervision, which is common to many fields as labelling data is costly. In particular, it is useful in the context of aviation because of the lack of labeled data. In this context, we can consider the bag to be a flight and the instances to be the time-steps of the features. Furthermore, positive bags are flights that had a safety event while negative bags are the ones with no events recorded. The standard MIL assumption, which is used in this work, states that all negative bags contain only negative instances, and that all positive bags contain at least one positive instance \cite{MIL}. This means a flight experiencing an event have at least one abnormal time-step that caused the whole flight to be considered a positive bag. Flights can later be divided into training, validation, and testing sets, and the performances of a chosen model can then be evaluated on those two subsets.

The learning task for this work is a classification problem. In a classification application, the task is to use given data and assign it to one of multiple predefined classes. In particular, given time-series from sensor data, we can assign the series a label such as a safety event \cite{survey_data_mining}. MIL classification task can be performed at the bag and instance level. However, the methodology developed in this work will use the bag label to infer the instance label of each feature at each time-step. In other words for example if we classify a bag as positive, the algorithm will determine (infer) the time instances that enabled the classification of the bag.

\section{Methodology}

\begin{figure}[hbt!]
 \centering % Always center your figures and tables.
 \includegraphics[width=0.8\linewidth]{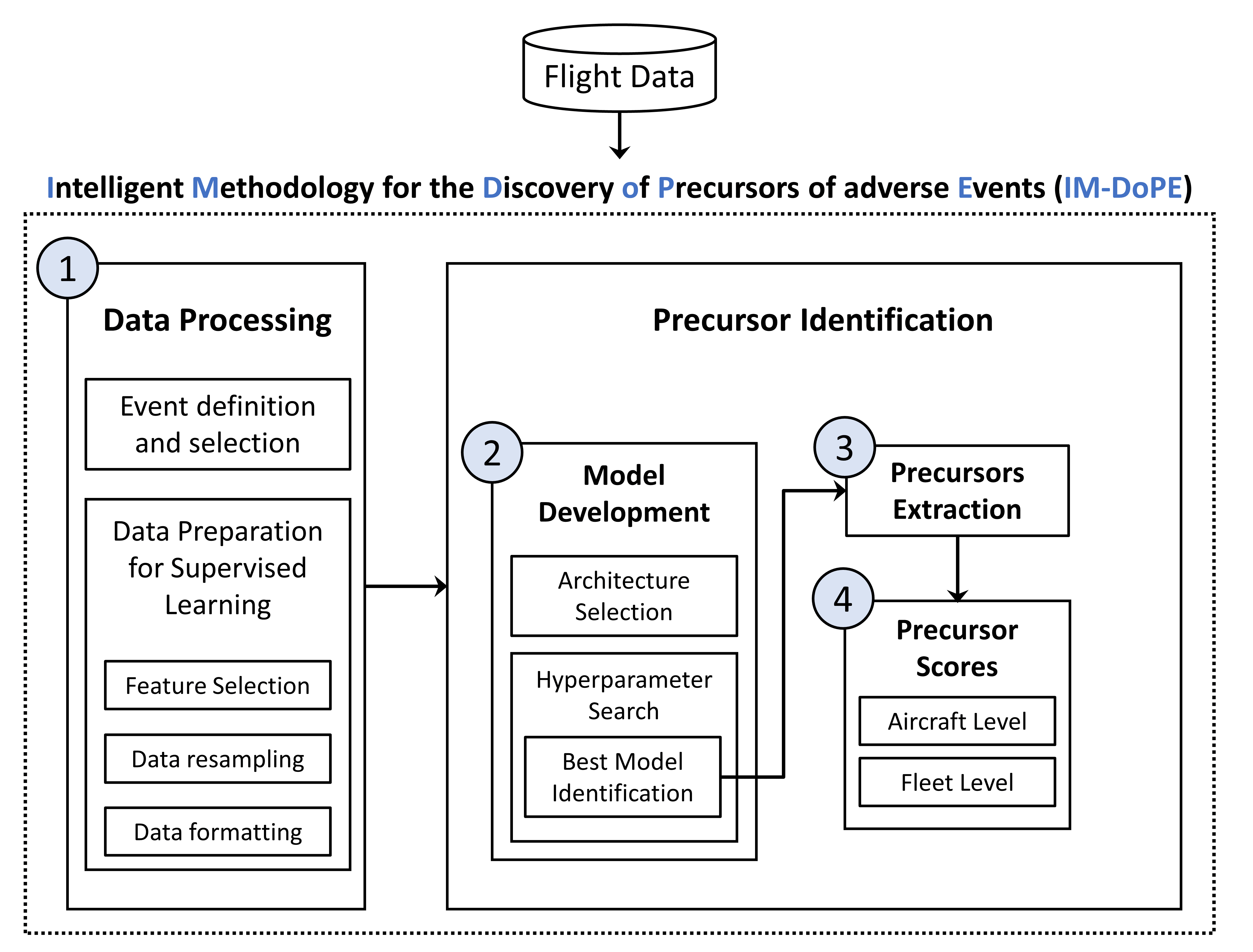}
 \caption{Intelligent Methodology for the Discovery of Precursors of Adverse Events}
 \label{fig:imdope}
\end{figure}

The Intelligent Methodology for the Discovery of Precursors of Adverse Events (IM-DoPE) developed in this work is described in this section. The four steps of the methodology as seen in Fig. \ref{fig:imdope} are the data processing, the development of the model, the extraction of precursors and the derivation of precursor scores. These steps allow for the discovery of precursors related to adverse events of interest, which can be used to provide insights into the cause of these events. 

\subsection{Data Processing}
The DASHlink sample flight data is recorded at 1 Hz frequency, therefore no effort was placed in re-sampling the data set at the same frequency, as this is a sufficiently high frequency. The data set had already been cleaned prior to accessing it as no missing data was observed. Subsequent tasks mainly focused on selecting events of interest, dimensionality reduction, data interpolation,  and data formatting.

\subsubsection{Event Definition and Selection}
\begin{figure}[hbt!]
 \centering % Always center your figures and tables.
 \includegraphics[width=0.58\linewidth]{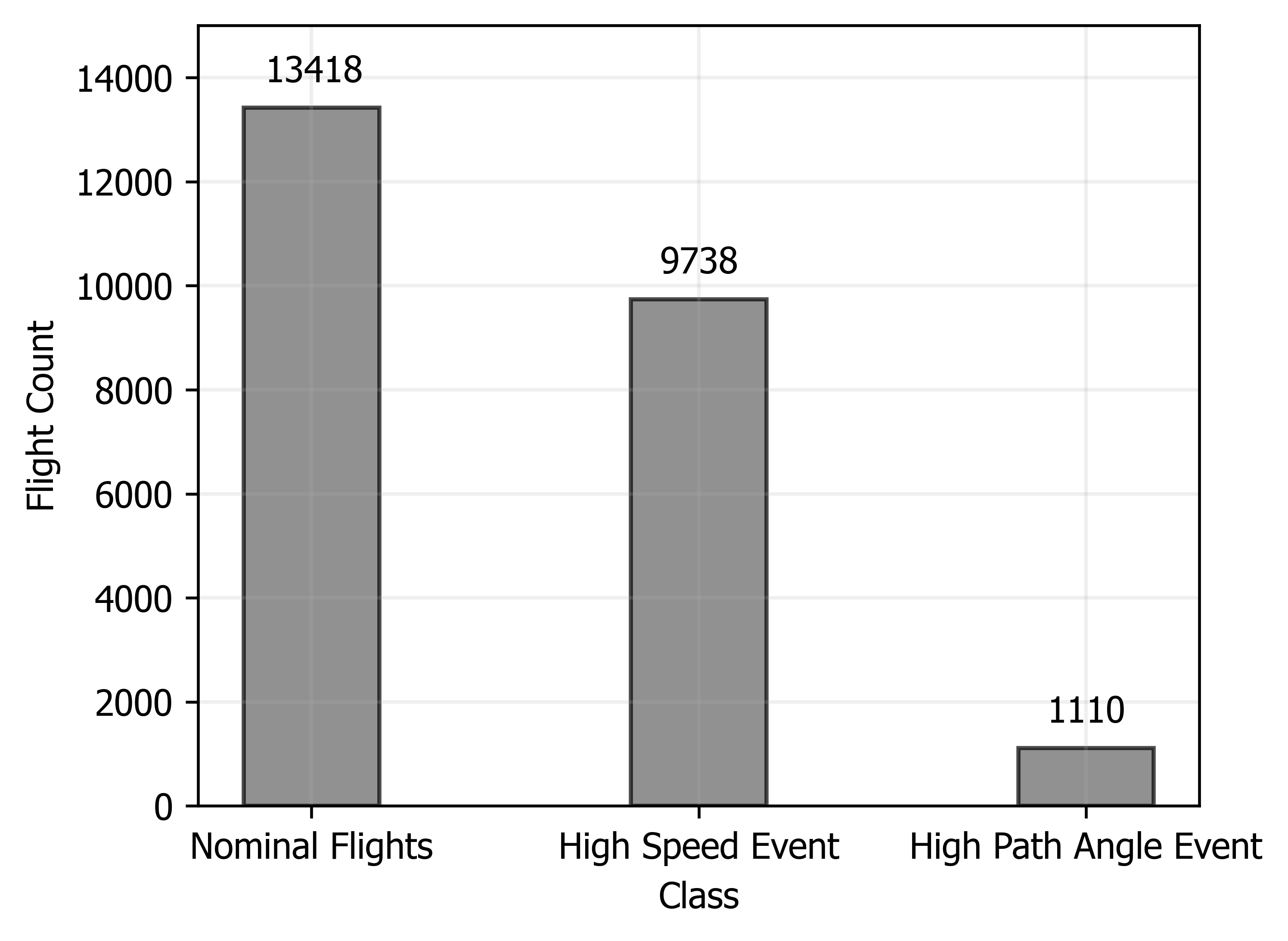}
 \caption{Flight Counts for Nominal and Adverse Flights}
 \label{fig:flcounts}
\end{figure}

As previously stated, 2 adverse events were selected for this work due to their higher frequency in the available data set. These events are pre-defined and characterized by the deviation of certain aircraft parameters from accepted nominal behaviors. Fig \ref{fig:flcounts}, shows the counts of nominal flights, and adverse flights that experienced one of the two events. As seen in the figure, the high speed event was the most commonly observed event among the available data followed by the high path angle event. 

\subsubsection{Feature Selection}
The dimensionality of the data was reduced by leveraging the correlations among the features. Certain parameters (e.g. computed airspeed and true airspeed) are different but convey similar information. This is due to the redundancy of parameters, the aircraft's physics, and the derivation of parameters \cite{precursor_jamey}. Using the data, a correlation matrix was created. For each pair of features, the Pearson's correlation coefficient is computed. The coefficient is a measure of the strength of the association between the features in the pair\footnote{Data Analysis - Pearson's Correlation Coefficient: \url{http://learntech.uwe.ac.uk/da/Default.aspx?pageid=1442}}. The Pearson correlation coefficient for two random variables $X$ and $Y$ is given by: 
\begin{equation}\label{eq:correlation}
    \rho = \frac{\sigma_{XY}^2}{\sqrt{\sigma_{XX}^2\sigma_{YY}^2}}
\end{equation}
The correlation coefficient $\rho$ is bounded between $-1 < \rho < 1$, and values closer to 1 signify strong positive correlation while values closer to -1 suggest strong negative correlation. In eq. \ref{eq:correlation}, $\sigma_{XX}^2$ is the sample variance for the variable $X$, and $\sigma_{YY}^2$ the one for the variable $Y$, and $\sigma_{XY}^2$ the co-variance between $X$ and $Y$. Since features are considered highly correlated if they have a correlation coefficient higher than $0.9$ \cite{corrcoef}, a threshold of $\rho=0.90$ was set so that for a given pair of features, with the absolute value of their correlation coefficient greater or equal to the threshold, one of the correlated features was removed. Note that the absolute value of the correlated coefficient between feature pairs was taken since since feature can be highly negatively correlated. For this work, only non highly correlated continuous variables were used as the input feature space. In addition to the correlation-based feature selection, trivial precursors were removed. This step was performed to avoid observing  one of the speed parameters as precursors to the high speed event. 

\subsubsection{Data resampling}

\begin{figure}[hbt!]
    \centering
	\includegraphics[width=.9\textwidth]{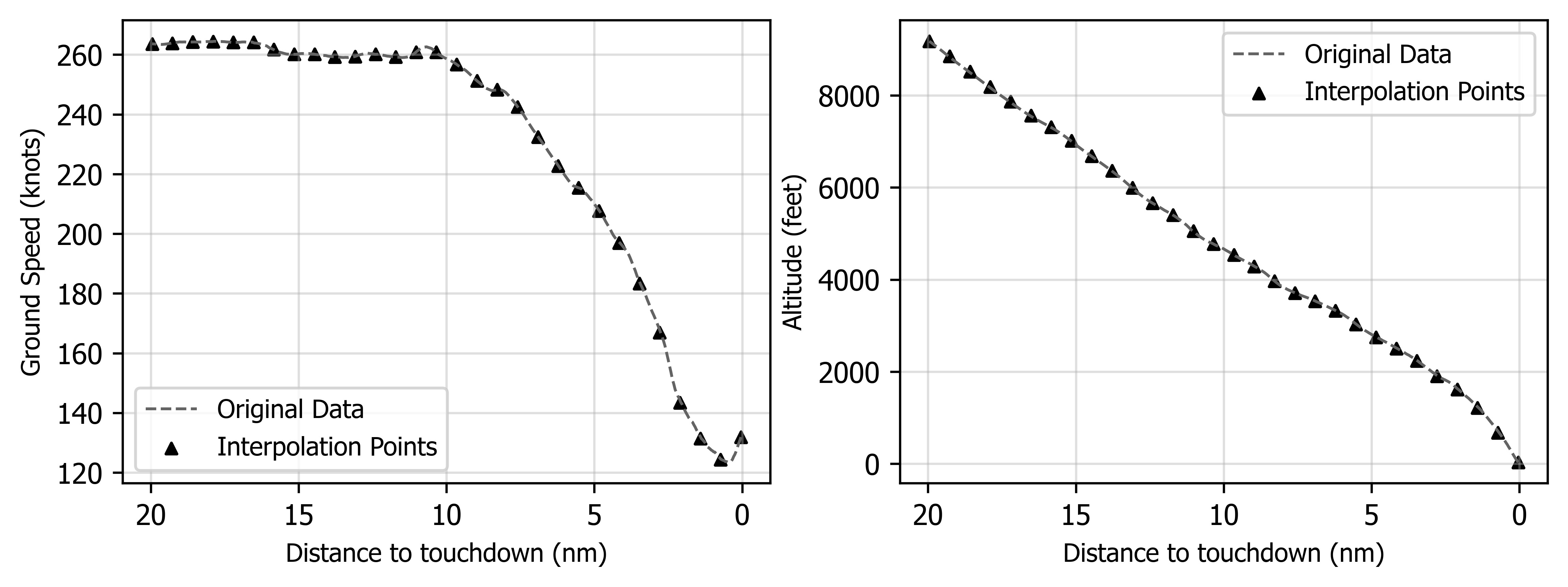}
	\caption{Example of Altitude and Ground Speed Interpolations}
	\label{fig: interpolation}
\end{figure}

All the events of interest for this work are flagged at a 1,000 ft above touch-down. Given the large number of flights and the differences between each of them, re-sampling the data was necessary to ensure uniformity across all flights. Thus, starting from a distance of 20 nautical miles away from the 1,000 ft mark, flights were re-sampled at every quarter nautical miles. Each flight therefore contains 81 data points or time steps. A Python code was developed to interpolate each feature to retrieve their values at every quarter mile distance, yielding a data set similar to table \ref{tab:resampled} for each flight. Fig. \ref{fig: interpolation} shows two examples of interpolated data. The original flight data contained 281 points for each feature while the interpolated data only contain 30 points.

\begin{table}[hbt!]
\caption{Example Results of Interpolating Parameters For a Given Flight}
\label{tab:resampled}
\begin{center}
\begin{tabular}{ccccc} \hline\hline
 \textbf{Distance away from 1,000 ft} & \textbf{Feature 1} &  \textbf{Feature 2} & \dots & \textbf{Feature d}\\\hline \hline
20 & X & X & \dots & X \\ 
19.75 & X & X & \dots & X \\ 
\vdots & X & X & \dots & X \\ 
0 &  X & X & \dots & X \\ \hline\hline
\end{tabular}
\end{center}
\end{table}

\subsubsection{Data Formatting}
\begin{figure}[ht]
    \centering
	\includegraphics[width=0.8\textwidth]{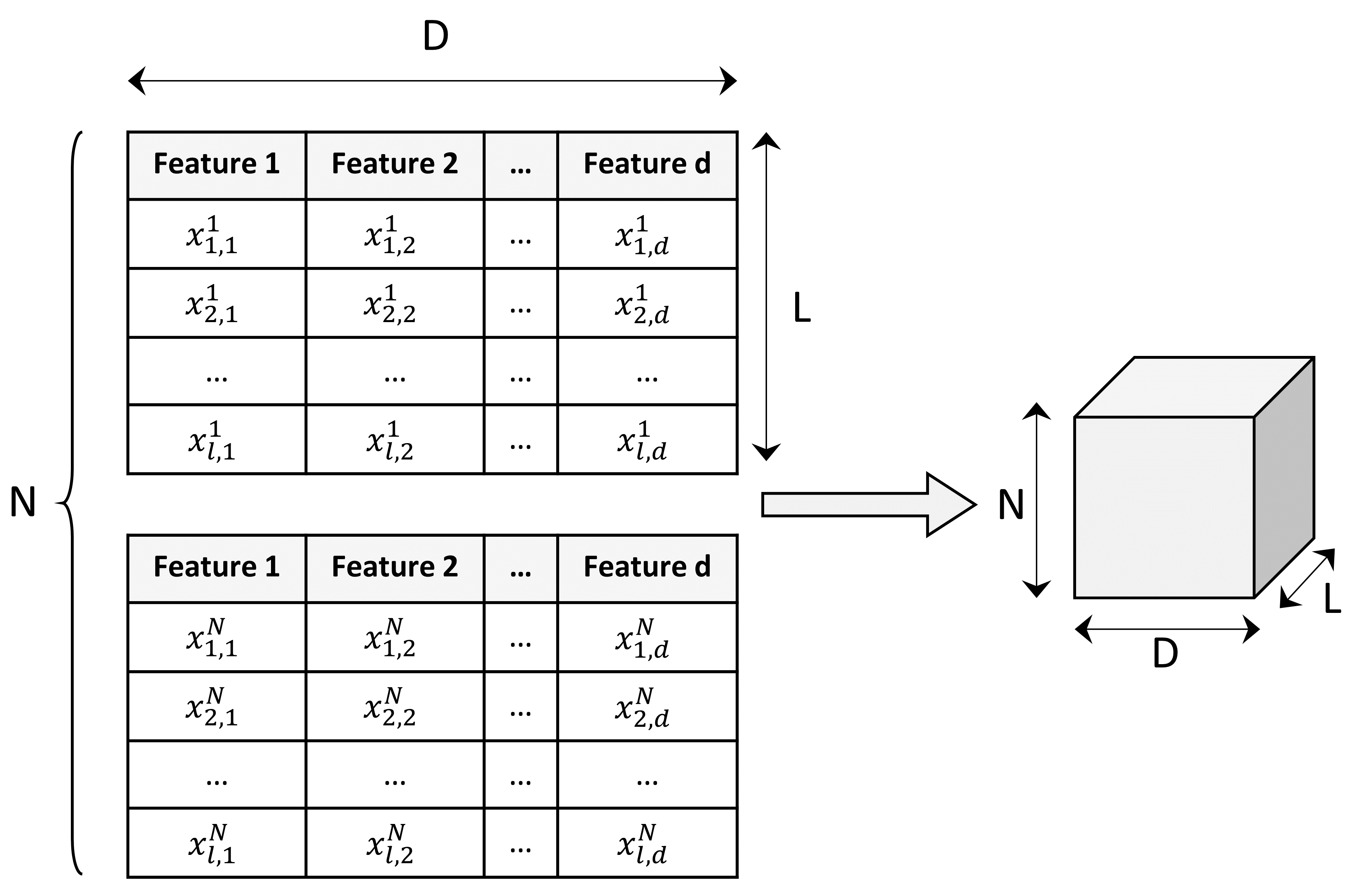}
	\caption{Flight Data Reshaping}
	\label{fig:tensor}
\end{figure}

In practice, modern algorithms such as Convolutional Neural Networks (CNNs), and Recurrent Neural Networks (RNNs) usually take as inputs 3 dimensional tensors. In order to account for this, data from each flight, as specified in table \ref{tab:resampled}, is concatenated together into a tensor as as shown on fig. \ref{fig:tensor}, with $X^f_{i,j}$ referring to feature $j$ for flight $f$ at time step $i$.

The data takes a 3 dimensional form where the first dimension (N) corresponds to the batch size or the number of flights that the deep learning algorithm will process at once. The second dimension (L) is the number of time-steps in each flight equal to 81 for this work. Finally the last dimension (D) refers to the number of features/aircraft parameters which corresponds to 58 continuous variables. 

\subsection{Model Architecture}\label{subsection:model architecture}
The model architecture was chosen to take advantage of the Multiple-Instance Learning (MIL) framework, the feature extraction capabilities of Convolutional Neural Networks (CNNs) \cite{multiheadedcnn, Goodfellow-et-al-2016}, and the temporal pattern recognition ability of Recurrent Neural Networks (RNNs) yielding a MHCNN-RNN architecture. This architecture choice allows the aircraft's parameters to be initially processed individually by the MHCNNs, which yields feature maps. The complex (time-dependent) correlations between the feature maps are then learned by the RNN. The model was developed using PyTorch \cite{NEURIPS2019_9015}, a high performance Python deep learning library. 
\subsubsection{Architecture Selection}\label{subsub:architecture selection}
% The MIL framework is useful because it allows the usage of the flight level label to correctly predict an adverse event and retrieve the time instance at which the used algorithm found a precursor to this event. More specifically,

\begin{figure}[hbt!]
 \centering % Always center your figures and tables.
 \includegraphics[width=0.85\linewidth]{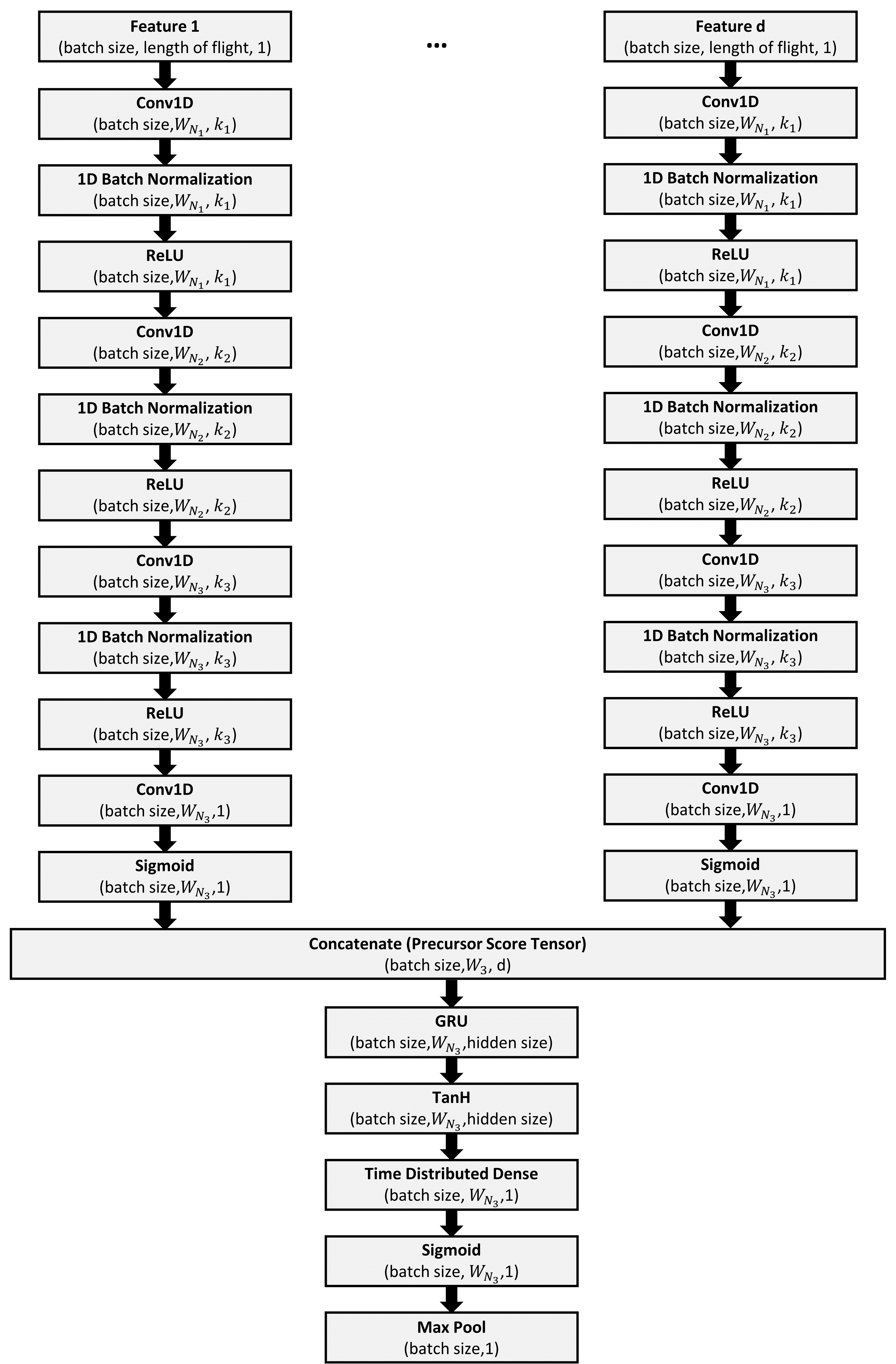}
 \caption{Layer configuration with output size of layer}
 \label{fig:architecture1}
\end{figure}

\begin{figure}[h!]
 \centering % Always center your figures and tables.
 \includegraphics[width=0.9\linewidth]{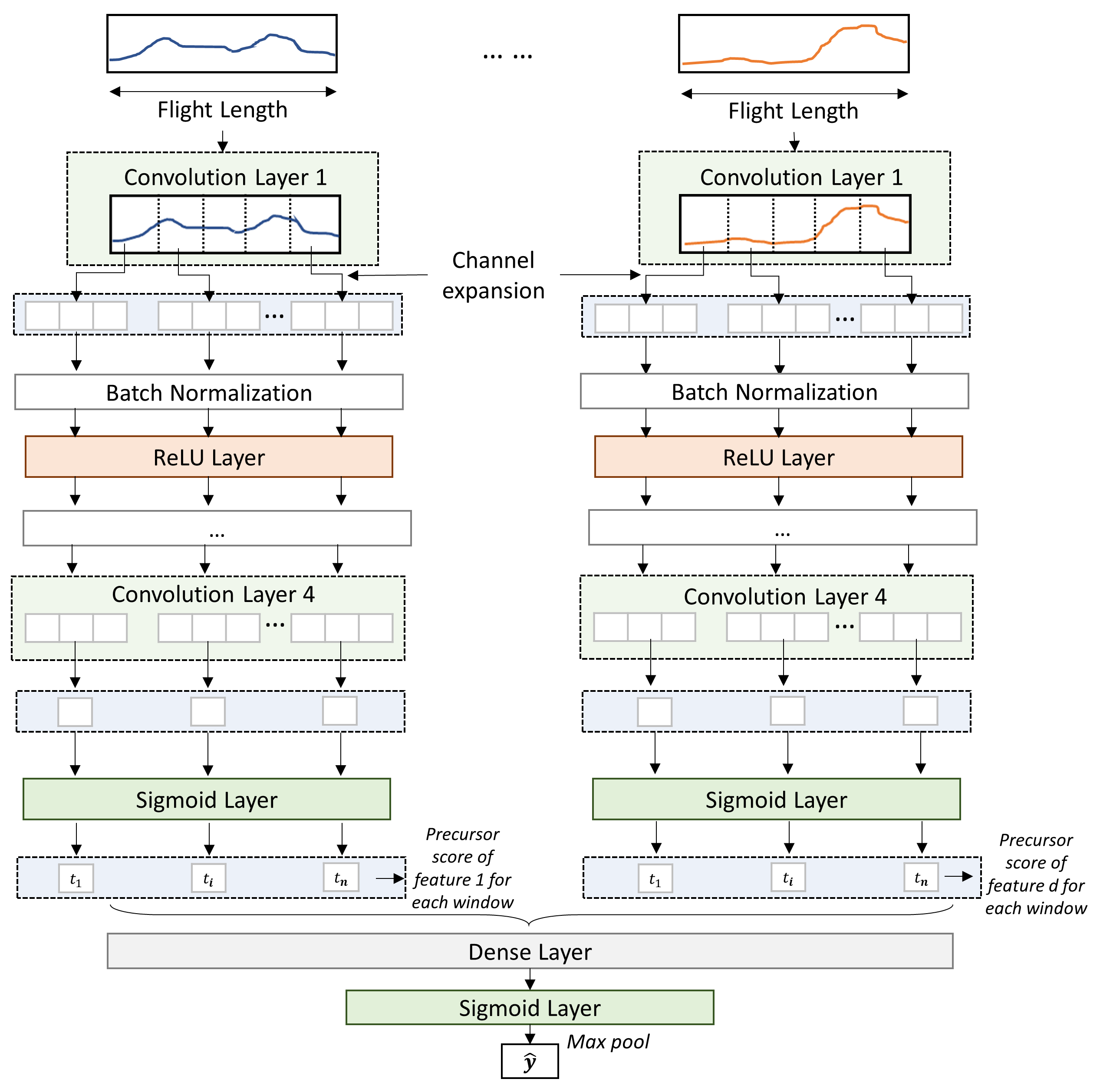}
 \caption{Layer configuration showing where the precursor score is extracted}
 \label{fig:architecture2}
\end{figure}

The MHCNN-RNN architecture combines a Multi-Head CNN (MHCNN) to a Gated Recurrent Unit (GRU), a type of RNN, as shown on fig. \ref{fig:architecture1}. The MHCNN is effective at processing each feature independently \cite{multiheadedcnn}. This allows the model to find relevant patterns in each aircraft parameter, which subsequently leads to the identification of the parameters that are highly correlated to the adverse event. Indeed, the most important aircraft parameters are the ones that helped the algorithm correctly predict an anomaly, and therefore can be considered precursors to the anomaly. A window-based approach was used to process the data with the MHCNN, allowing for information extraction of each feature at different time regions. The window is defined by the kernel size of the CNN and the networks all use a stride of 1, such that overlapping windows are created. Some instances of time are more useful than others to predict an anomaly and this approach allows the algorithm to have a more granular extraction of features. Similar to the architecture proposed in \cite{multiheadedcnn}, four CNNs are used in each head to process the time-series and expand the unique channel of each time series as seen in fig \ref{fig:architecture1} and fig \ref{fig:architecture2}, resulting in the extraction of relevant information from each window of the time-series. In Figure \ref{fig:architecture1}, $k_1, k_2 \text{ and } k_3$ represent the number of channels, and $W_{N_1}, W_{N_2}, \text{ and } W_{N_3}$ represent the length of the time series after each convolution operation. The length of the time series after each operation depends on CNN parameters such as the kernel size, the stride, and whether or not padding was used.The first three convolutional layers use batch normalization to reduces the internal co-variance shift, and bring a regularization effect \cite{multiheadedcnn}, and each normalization is followed by a ReLU activation function which has become a default function to use when developing neural networks due to its many advantages \cite{MLmystery_relu}. The fourth convolutional layer reduces the number of channels back to 1, and applies a sigmoid activation function, allowing the output (or feature map) to be interpreted as the probability of a given feature to be a precursor. The chronological order of the feature maps was kept, which helped maintain the temporal information of the data. A concatenation step was then performed to combine the feature maps of each aircraft parameters into a 3 dimensional precursor score tensor.

The concatenated tensor is the input to the GRU, which processes the features all together unlike the MHCNN that processed the features independently. This layer learns the temporal patterns present in the extracted information by the previous convolutional layers. The GRU is appropriate for handling temporal data due to its structure characterized by two gates: the update and reset gates \cite{Goodfellow-et-al-2016}. Both of these gates are used to select which information to keep and which to ignore, and allow the GRU to handle long-term dependencies better than simpler architectures \cite{Goodfellow-et-al-2016}. An hyperbolic tangent activation is then applied to the output of the GRU and followed by a time distributed dense layer that squeezes all the features into one unique dimension, allowing for additional approximation capability \cite{precursor_vijay}. The output is passed through a sigmoid function, and can be interpreted as the probability that a precursor occurs at a given instances of time. Finally to classify a given flight as positive (event occurred), it is assumed that if any instance of time is positive then the whole flight would be labeled as positive (MIL assumption). Thus, a max pooling layer is used to get the maximum probability across time to classify the flight. A threshold is set such that a flight is positive only if the maximum probability across time is greater or equal to the threshold. The architecture is thus expected to achieve two critical goals:
\begin{enumerate}
    \item Label the bag for identifying adverse events
    \item Extract precursors and the time instance at which they occur
\end{enumerate}

% The second convolutional layer takes in all the windows and sub-features and summarizes the information extracted by outputting a single channel for each window. The output is passed through a sigmoid function to bound the values between 0 and 1, which can be interpreted as the probability of being a precursor. Therefore, the output of this layer yields the probability of a feature being a precursor for each of the $W_N$ windows. The probabilities of each feature and every window are then concatenated into a tensor which is sent to a time-distributed fully connected dense layer which combines all the sensors together to retrieve the time instance at which the precursor occurred. Figure \ref{fig:architecture2} shows the layer, right before the dense layer, that is used to extract time-instance probabilities of each feature from the model. Similarly as before, the output is passed through a sigmoid layer to generate a probability across time, which is then max-pooled such that the largest probability at a given time instance is used to label the flight (bag-label). Note that the probability across time are extracted from the model in order to determine the time region where the model flagged an increase in the probability of seeing an adverse event.

Although the two approaches are similar because they take advantage of the MIL framework for precursor mining, it is important to note the differences between this architecture and the previously developed algorithm ADOPT \cite{precursor_vijay}. The architecture proposed in this paper is designed such that the precursor probabilities for each feature can be extracted directly from the neural network layers, this approach aims at creating a more interpretable model. ADOPT can only identify temporal precursors directly, and needs to perform a sensitivity analysis during a post-processing step, which consist of perturbing one individual feature at a time and measuring its effect on the time instance probabilities to determine precursors in the feature space. This method potentially misses interactions between features. Thus being able to retrieve precursors by leveraging the combination of the MHCNNs (extraction of individual information for each parameter) and the GRU (extraction of temporal interactions of parameters) mitigates this drawback while eliminating the overhead required to setup the post-processing analysis.

\subsubsection{Multi-Class Extension}\label{subsub:multiclass}
\paragraph{Multiple Binary Classifiers}
The architecture described in \ref{subsub:architecture selection} corresponds to a binary classifier. Both IM-DoPE and ADOPT make use of binary classifiers as their task is not only to correctly predict the occurrence of an adverse event but to also identify precursors to the event. Both models could be extended to include multiple outputs allowing for the training of multiple events at once \cite{precursor_vijay}. However, for this work the multi-class model is obtained by combining multiple binary classifiers since it was observed that they performed well at completing their individual tasks. Two popular approaches of combining the classifiers are the "one-vs-all" and the "one-vs-one" strategies \cite{GALAR20111761}. The first strategy learns to classify one class at a time, where the class is distinguished from the other ones. The second strategy requires creating multiple classifiers for different pairs of classes. In this work, the approach taken is a mix between the two strategies. Indeed, 2 classifiers are trained for each event, similarly to "one-vs-all" approach, but each of them are trained to either recognize the event or the default nominal operation. This approach was chosen to allow for the interpretation of nominal flights when performing predictions and to avoid training extensive number of models. When using the models to perform an inference, the output class is chosen to be the output of the model with the highest probability (i.e. the highest confidence) that is greater than a specified threshold. If none of the probabilities are greater than the threshold, then the output is defaulted to the class corresponding to normal operations since each model either predicts an event or normal operation. When using sigmoid functions, the decision threshold is problem dependent and could be treated as a hyperparameter to tune. However, a default decision threshold of 0.5 was used for this work since it yielded good performances for the model. The advantages of this architecture are the clear interpretability of the probability of a flight experiencing a given event or not, and the easier learning task that the model has to accomplish. 

\paragraph{Multiple Output Nodes}

The binary architecture can also be changed to include multiple output nodes, such that the output after the last max pool layer is of size $(batch size, c)$ where $c$ is the number of classes to predict. The sigmoid layer before the final max pooling was kept to ensure that for each class/anomaly, the probability that the flight experiences each of them is bounded between 0 and 1. Indeed this allows for multi-labeling, which is a use case in aviation since multiple events could occur during the same flight. Using a sigmoid layer is different than using a softmax layer, which would require the sum of the probabilities of each event to be 1. This architecture was preferred over the usage of a softmax layer because it could handle both multi-class, and multi-label problems. In fact for this work the model was trained using flights that experienced only one anomaly at a time, and therefore learned to keep the probabilities of non-occurring events low while keeping the ones of occurring events higher. Thus, for each flight the class with the highest probability was chosen as the final output class. The advantage of this architecture is the convenience of being able to train one model instead of multiple binary classifiers.

\subsubsection{Hyperparameter Search}
Given the novelty of the chosen architecture for the precursor mining task, there is a lack of knowledge on what parameters to choose for the model, and on how to train it. To mitigate this problem, a hyperparameter search was performed to determine the optimal parameters of the architecture for each event. A grid-search is the most straight-forward strategy to perform a hyperparameter search, as it entails searching through the space created by all possible combinations of hyperparameters. This means that each parameter is given an array of values to try, and the model is evaluated using metrics described in \ref{subsub:metrics} for each of the parameters combinations. Table \ref{tab:hyperparam} defines the hyperparameter space containing 36 possible combinations.

\begin{table}[hbt!]
\caption{Hyperparaneters and Search Space}
\label{tab:hyperparam}
\begin{center}
\begin{tabular}{cc} \hline \hline
 \textbf{Hyperparameter} & \textbf{Search Range} \\
\hline \hline
Convolutional Layers 1,2,3 kernel sizes & [8, 5, 3],
           [6, 3, 2] \\ 
Convolutional Layers 1,2,3 output channels & [10, 15, 20], 
[16, 32, 64],
[32, 64, 128] \\ 
Learning Rate & 0.001, 0.0001 \\ 
Weight Decay (L2 regularization) &  0.01, 0.001, 0.0001  \\
Mini-batch size percent & 1\% of training set size \\ 
Optimizer & ADAM \cite{Kingma2015AdamAM}\\
Number of epochs & 30 \\ \hline \hline
\end{tabular}
\end{center}
\end{table}

\begin{figure}[ht]
    \centering
	\includegraphics[width=.7\textwidth]{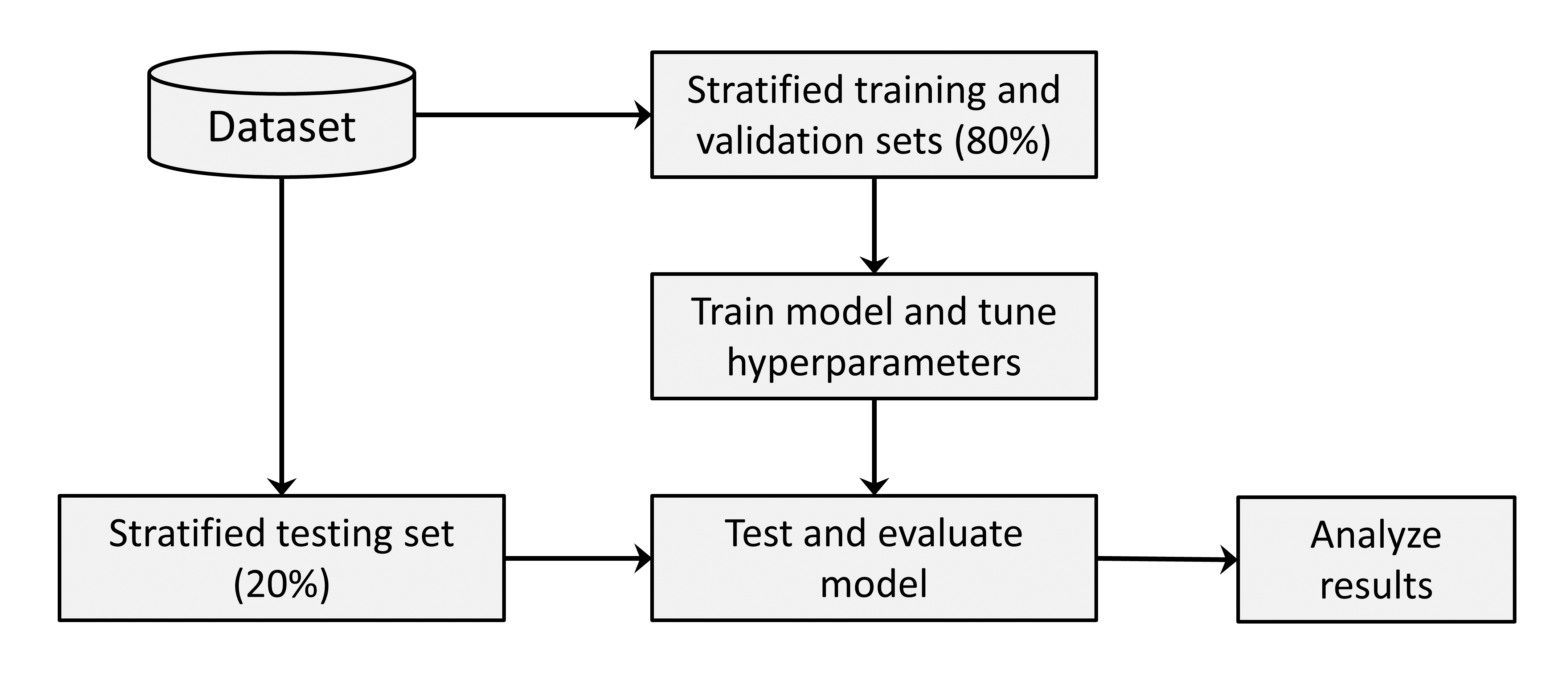}
	\caption{Data Split for Model Training, Hyperparameter Tuning, and Model Evaluation}
	\label{fig:datasplit}
\end{figure}

Similarly to \cite{ml_gdp_gs_coincidence}, the data set was divided into 3 subsets: the training set, the validation set, and the testing set as seen on fig. \ref{fig:datasplit}. Stratified subsets are used to ensure that the proportion of nominal flights to abnormal flights remain consistent in each subset. Additionally, stratified mini-batch were used during the training process. The training set was used to train the model using a given set of hyperparameters (i.e. one hyperparameter combination). Google Colab's Tesla V100 16 GB of RAM GPU was leveraged to train the models. The performances of the models on the validation set was used to determine the best model. 

\subsubsection{Classification Metrics}\label{subsub:metrics}
Several metrics are used to quantitatively evaluate the model's performance. The evaluation allows for the selection of the best model obtained from the hyperparameter search and its assessment on new unseen data (i.e. on the test set). The best model correctly forecast the occurrence of an adverse event and therefore is best suited to determine the aircraft's parameters that are the most correlated to the event. The following metrics were used for the model selection and evaluation:
\begin{enumerate}
    \item \textbf{Confusion Matrix}: Contains the counts for true positives, true negatives, false positives and false negatives \cite{ML}, as shown by: \begin{table}[hbt!]
\caption{Confusion Matrix}
\begin{center}
\begin{tabular}{p{5cm}p{5cm}p{5cm}} \hline\hline
 & \textbf{Predicted: No Event}  & \textbf{Predicted: Event}
 \\
\hline\hline
\textbf{Actual: No Event}
 & True Negative (TN) & False Positive (FP)
  \\ 
\textbf{Actual: Event}
 & False Negative (FN) & True Positive (TP)
   \\ 
\hline\hline

\end{tabular}
\label{CMa}
\end{center}
\end{table}
    \item \textbf{Precision}: Measures the proportion of positively labeled examples that are actually positive \cite{MLBOOK}. The closer to one the better, the following equation defines the precision: 
    \begin{equation}
        \text{precision}= \frac{TP}{TP+FP}
    \end{equation}
    \item \textbf{Recall}: Measures the the fraction of positives labels that were actually detected \cite{MLBOOK}. The closer to one the better, the following equation defines the recall:
    \begin{equation}
        \text{recall} = \frac{TP}{TP+FN}
    \end{equation}
    \item \textbf{F1 score}: Harmonic mean of the the precision and recall \cite{MLBOOK}. The closer to one the better, the following equation defines the recall:
    \begin{equation}
        \text{F1 score}= \frac{2\times TP}{2\times TP+FN+FP}
    \end{equation}
    \item \textbf{Distance From ADOPT (DFA)}: Metric created to measure the resemblance in the precursor score ranking of IM-DoPE and ADOPT \cite{precursor_vijay2}. Assuming precursor scores $p_i$ for feature i ranked between 0 and 1, N flights, and d feature, the DFA for each combination of hyperparameters is given by:
    \begin{equation} \label{sse}
    sse_{j} =  \frac{1}{d} \sum^d_i \left( p_{i,IMDoPE}- p_{i, ADOPT} \right)^2 
\end{equation}
    \begin{equation}\label{dfa}
        \text{DFA} = \frac{1}{N}\sum^{N}_{j} sse_{j} 
    \end{equation}
    DFA for a combination of hyperparameters is the mean sum square error between the precursor scores of ADOPT and IM-DoPE, therefore the closer to zero, the more alike the rankings of the two algorithms are. The precursor score sum square errors are averaged across all features \ref{sse}, and across all flights \ref{dfa}. ADOPT has been validated by expert for few events such as an high speed exceedence and can thus be used as a second-hand validation if no human expert is available. 
 \end{enumerate} 
 
\subsection{Precursor Identification}
\subsubsection{Raw Precursor Scores}
\begin{figure}[ht]
    \centering
	\includegraphics[width=.99\textwidth]{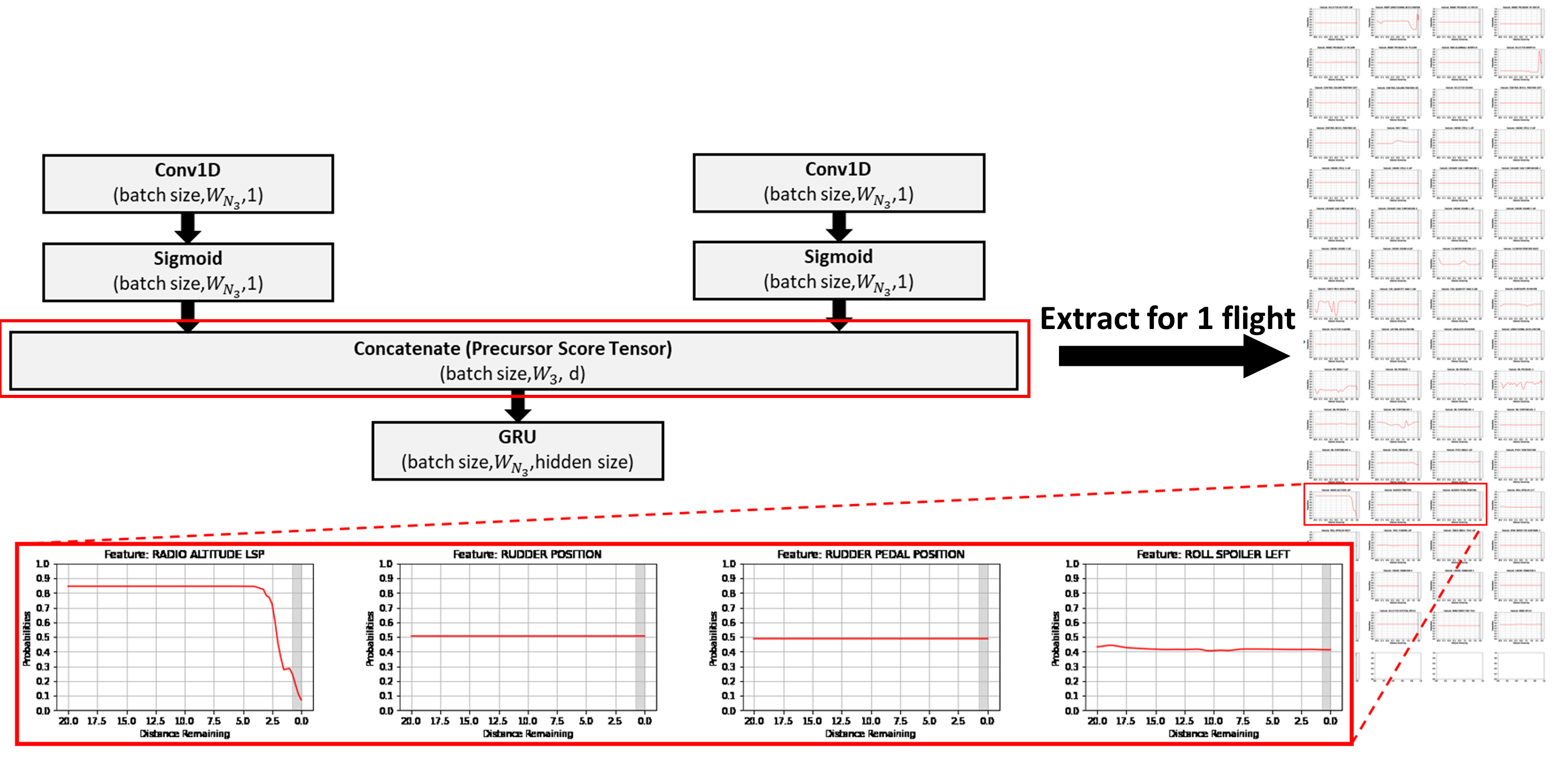}
	\caption{Extraction of Precursors for a High Speed Event}
	\label{fig:p_out}
\end{figure}

Once acceptable performances are achieved by the model, it can be used to identify precursors. The precursor score $p_i,t$ of each feature $i$ at time step $t$ is extracted directly from the architecture. As seen on fig. \ref{fig:p_out}, the raw precursor score for each feature can be extracted from the last convolution layer of each head. The raw scores are bounded between 0 and 1 due to the sigmoid function used in the fourth convolulational layer. From empirical results, it was observed that the model learns to set non-important parameters to $0.5$. As an example on Fig. \ref{fig:p_out}, the rudder position, the rudder pedal, and the spoiler position are deemed not important by the model since their individual scores remain at $0.5$ for last 20 nautical miles before a 1,000 ft. On the other hand, the radio altitude was flagged as a precursor. As the less important features go through the different convolutional layers, they get reduced to zero. This zeroed out feature is then passed through the sigmoid activation resulting in a raw score of $0.5$. 

\begin{figure}[ht]
    \centering
	\includegraphics[width=.85\textwidth]{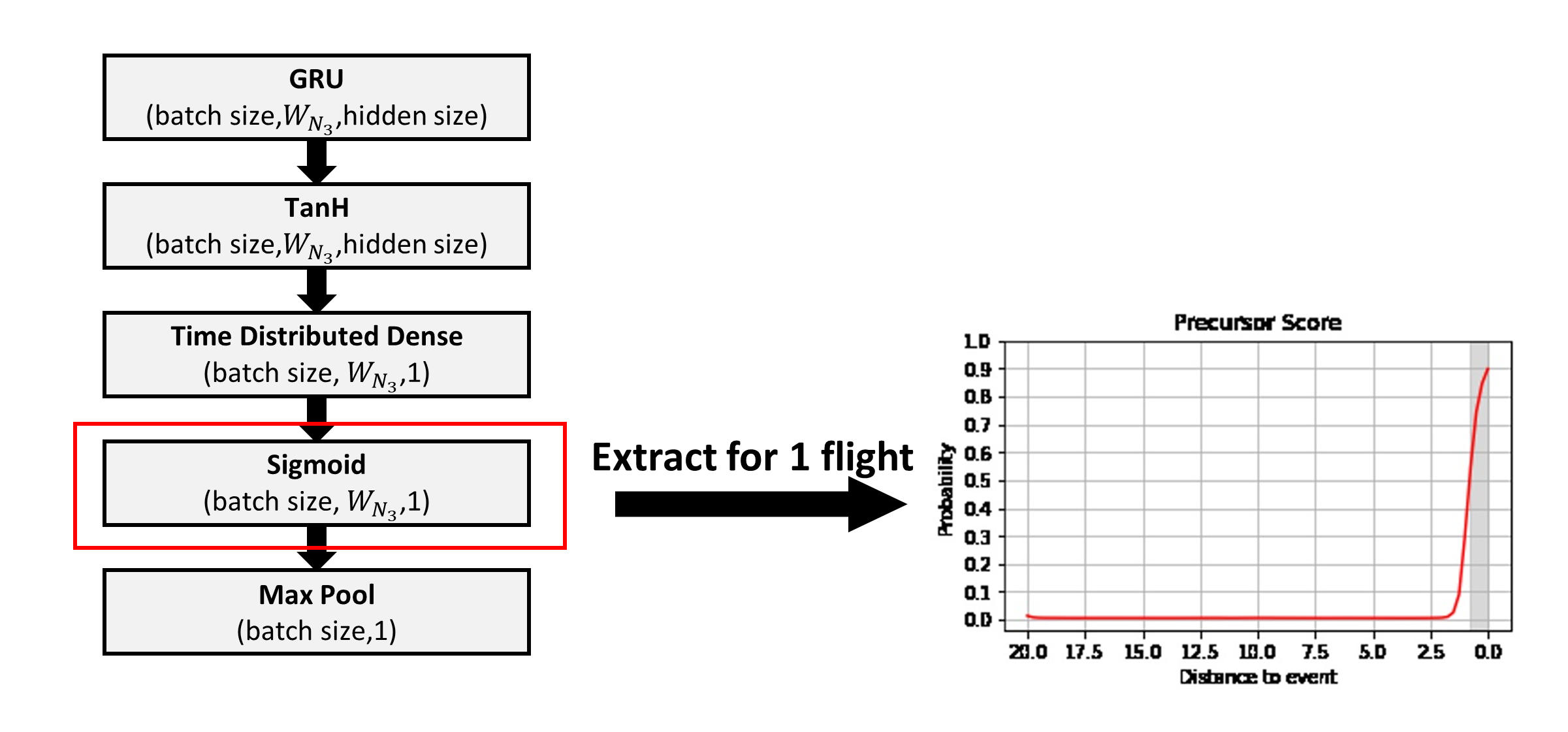}
	\caption{Extraction of Precursors Score Over Time from Dense Layer for a High Speed Event}
	\label{fig:p_temp_out}
\end{figure}

The identified precursors have instance of time (precursor score over time) at which they are more correlated to the event of interest (grayed area in Fig. \ref{fig:p_out} and \ref{fig:p_temp_out}). The combination of GRU and dense layer allows for the extraction of these time instances, as seen in Fig. \ref{fig:p_temp_out}. Similar to ADOPT, abnormal flights will have the precursor score increase towards 1 and nominal flights will have the score fall towards 0.  Additionally, since the precursors are more correlated to an event at time windows identified by the GRU, an overall score can be given to the precursor by averaging its $p_{i,t}$ values within an identified window $T$. 

\subsubsection{Adjusted Precursor Scores}\label{subsub: adj precursor score}
A more natural way to express the precursor score is to have a score of zero for parameters that are not important. Therefore, the raw precursor scores were adjusted.
The adjusted precursor score $p_{i}$ of feature $i$ is therefore defined by the following equation: 
\begin{equation} \label{eq:pprime}
    p_{i} =  \frac{1}{|T|}  \sum_{t} \left(| p_{i,t} - 0.5| \right) , \forall t \in T \text{ where } T \subseteq L 
\end{equation}
where $|T|$ is the number of time steps within the time  window $T$. The adjusted precursor score is thus bounded between $0$ and $0.5$. From this definition, it can be seen that features with higher scores are flagged as precursors.
% \FloatBarrier

\section{Results and Discussion}
\subsection{Quantitative Results}
\paragraph{Multiple Binary Classifiers}
For each event, 36 combination of hyperparameters were evaluated and the best classifiers were selected. Table \ref{tab:results} summarises the quantitative results obtained when evaluating the models on the test set. 

\begin{table}[hbt!]
\caption{Model Evaluation Results (Binary Classifications  with Multiple Binary Classifiers)}
\label{tab:results}
\begin{center}
\begin{tabular}{cccccc} \hline \hline
 \textbf{Event} & \textbf{Algorithm} & \textbf{F1 Score} & \textbf{Precision} & \textbf{Recall} & \textbf{DFA}\\
\hline \hline
High Speed & IM-DoPE & 0.90 & 0.84 & 0.97 & 0.0392\\ 
High Path Angle & IM-DoPE & 0.83 & 0.81 & 0.87 & 0.0152\\ \hline 
High Speed & ADOPT & 0.88 & 0.90 & 0.86 & N/A\\ 
High Path Angle & ADOPT & 0.70 & 0.56 & 0.90 & N/A\\ \hline \hline
\end{tabular}
\end{center}
\end{table}

\begin{figure}[ht]
    \centering
	\includegraphics[width=.99\textwidth]{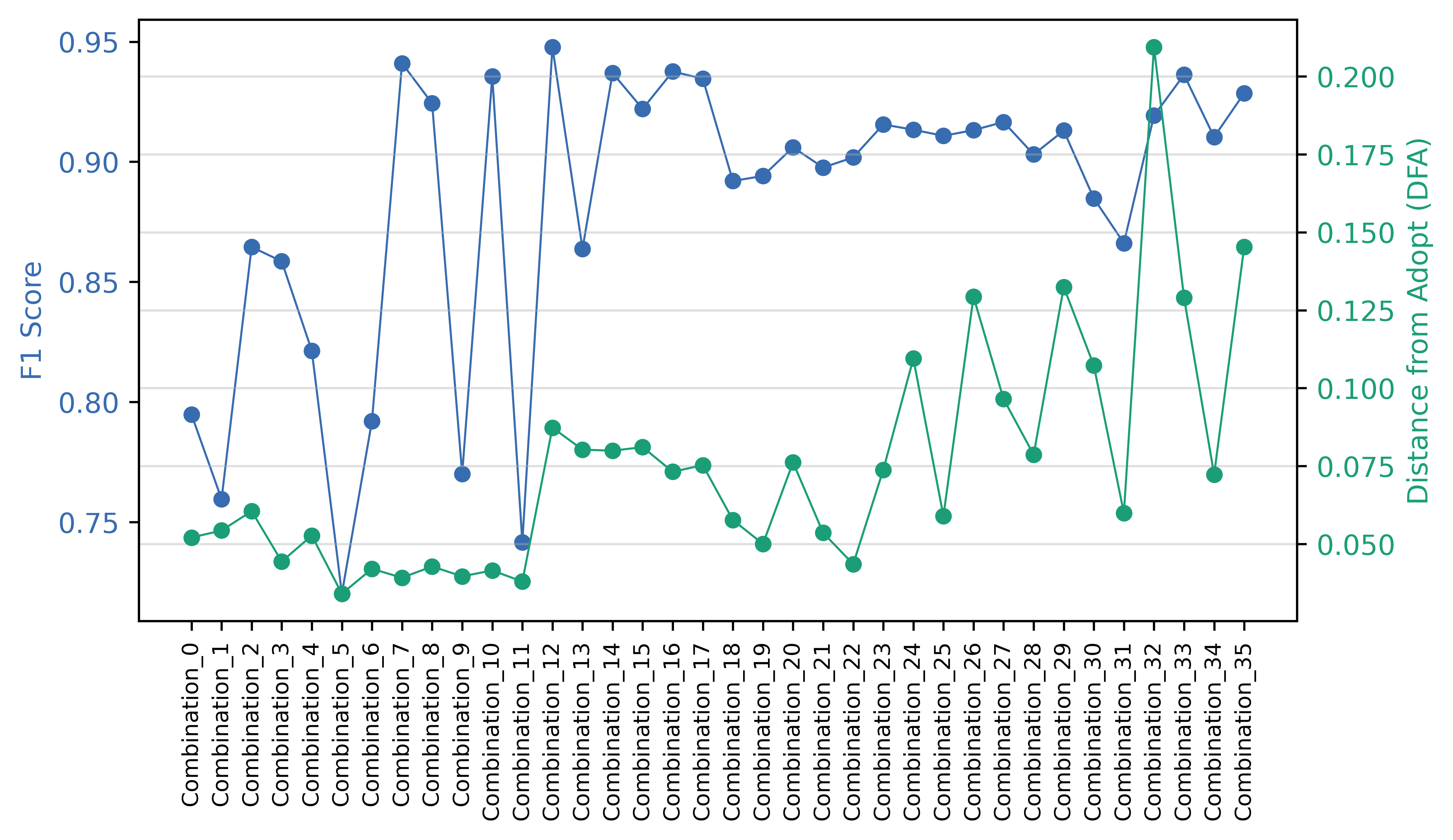}
	\caption{F1 Score and DFA for All Combinations (High Speed Event)}
	\label{fig:combi_f1_dfa}
\end{figure}

While the best MHCNN-RNN model achieves high performances, better results for the classical classification metrics are obtained for the high speed event due to the higher number of training examples and the the smaller imbalance in the classes. However, the larger DFA metric for the high speed event suggest a greater difference with ADOPT feature ranking for this event. Note that the DFA values obtained are still relatively good as it is among the lowest DFA values obtained as seen on fig. \ref{fig:combi_f1_dfa}. 
Additionally, best model of IM-DoPE performs better than ADOPT when using the same training set, and testing set. It is however important to note that while a grid-search was performed for IM-DoPE for each event, the default setting of ADOPT were used. Some models obtained during the grid-search have lower scores that are worse than the ones obtained from using ADOPT default settings. For example, on fig \ref{fig:combi_f1_dfa}, several models reached scores lower than 0.88. 

The binary classifiers are then combined as explained in \ref{subsub:multiclass} in order to obtain a multi-class classifier. Similar to binary classification, each metric can be computed for each class. For instance, the precision of the high speed class is  the number of flights that actually faced a high speed event over the total number of flight that were labeled as a high speed event by the classifier. Table \ref{tab:results_mc} presents similar scores as in table \ref{tab:results} showing that even when combined, each model learns to identify its positive class correctly. The confusion matrix in \ref{tab:CM} confirms this behavior, as the numbers of FP and TN for each class are relatively low.

\begin{table}[hbt!]
\caption{Model Evaluation Results (Multi-Class Classifications with Multiple Binary Classifiers)}
\label{tab:results_mc}
\begin{center}
\begin{tabular}{cccc} \hline \hline
 \textbf{Event} & \textbf{F1 Score} & \textbf{Precision} & \textbf{Recall}\\
\hline \hline
Nominal &  0.91 & 0.97 & 0.85 \\
High Speed &  0.90 & 0.84 & 0.97 \\ 
High Path Angle &  0.82 & 0.80 & 0.83 \\ \hline \hline
\end{tabular}
\end{center}
\end{table}

\begin{table}[hbt!]
\caption{Confusion Matrix (Multi-Class Classifications with Multiple Binary Classifiers)}
\label{tab:CM}
\begin{center}
\begin{tabular}{cccc} \hline \hline
& \textbf{Predicted Nominal} & \textbf{Predicted High Speed} & \textbf{Predicted High Path Angle}\\ \hline\hline
\textbf{Actual Nominal} &  1717 (47\%) & 263 (7.2\%) & 33 (0.9\%) \\
\textbf{Actual High Speed} &  38 (1.0\%) & 1422 (39\%) & 1 ($2.7 \times 10^{-4} \%$) \\ 
\textbf{Actual High Path Angle} &  16 (0.44\%) & 12 (0.33\%) & 138 (3.8\%) \\ \hline \hline
\end{tabular}
\end{center}
\end{table}

\paragraph{Multiple Output Nodes}
Similarly, the multiple output nodes architecture was trained using the same list of hyperparameters. After training, the best set of hyperparameters can be chosen using the validation set and the final performances of the model on the unseen test set can be evaluated. The results of this evaluation are presented in table \ref{tab:results_mc_mon}. On one hand from the results obtained, it can be seen that there are slight improvements in the F1 score of the nominal and high speed classes thanks to the respective increase in the recall and precision scores. On the other hand, the ability to accurately predict the high path angle anomalies decreased. In fact, this model architecture is able to find most flights that experienced the high path angle event, but has a higher number of false positive than the other multi-class model leading to a much lower F1 score. 

\begin{table}[hbt!]
\caption{Model Evaluation Results (Multi-Class Classifications with Multiple Output Nodes)}
\label{tab:results_mc_mon}
\begin{center}
\begin{tabular}{cccc} \hline \hline
 \textbf{Event} & \textbf{F1 Score} & \textbf{Precision} & \textbf{Recall}\\ \hline \hline
Nominal &  0.93 & 0.93 & 0.92 \\
High Speed &  0.92 & 0.95 & 0.89 \\ 
High Path Angle & 0.69 & 0.55 & 0.92 \\ \hline \hline
\end{tabular}
\end{center}
\end{table}

\subsection{Identified Precursors}
Even though both the multiple binary models and the multiple output nodes model can be used along with the precursor identification part of the IM-DoPE methodology, the combination of multiple binary classifiers yielded better results, and therefore was chosen to perform the identification of precursors. The results are presented in the following subsections.
\subsubsection{Average Adjusted Precursor Scores}
As previously mentioned, the final model is expected to  predict adverse events and identify their precursors. The adjusted precursor score can be obtained for each feature and each flight. Considering true positive flights only, the average adjusted precursor score can be determined and the precursors can be identified on a fleet level, as seen in table \ref{tab:precursor_score} for the top 5 precursors of each event. Noticeable differences are observed in the precursor rankings for the two events. For example the glideslope deviation and the pitch angle are characteristics of a high path angle event while the N1 target relates to engine power which can be related to speed. Some resemblances are also observed, for instance the altitude is seen to be the precursor for both events. This is expected since the altitude above touch-down is used to define both events.
The trained model can also be used to analyze individual flights. Two flights experiencing a high speed event, and a high path angle event respectively, are analyzed \ref{subsub:hs} and in \ref{subsub:hp}.

\begin{table}[hbt!]
\caption{Average Adjusted Precursor Scores for High Speed and High Path Angle Events}
\label{tab:precursor_score}
\begin{center}
\begin{tabular}{c|cc} \hline \hline
\multicolumn{1}{c|}{\textbf{Precursor Rank}} & \multicolumn{2}{c}{\textbf{Average Adjusted Precursor Score}}  \\
 & High Speed Event & High Path Angle Event\\ \hline\hline
\#1 & Altitude: 0.31 &  Radio Altitude:0.31 \\
\#2 & Radio Altitude : 0.28  & Glideslope Deviation: 0.24 \\ 
\#3 & N1 Target: 0.25 & Pitch Angle: 0.21 \\ 
\#4 & Body Longitudinal Acceleration: 0.23 & Altitude: 0.18\\
\#5 & Lateral Acceleration: 0.14 & Flight Path Acceleration: 0.12\\
 \hline\hline
\end{tabular}
\end{center}
\end{table}

\subsubsection{High Speed Event Flight Analysis}\label{subsub:hs}
\begin{figure}[ht]
    \centering
	\includegraphics[width=.65\textwidth]{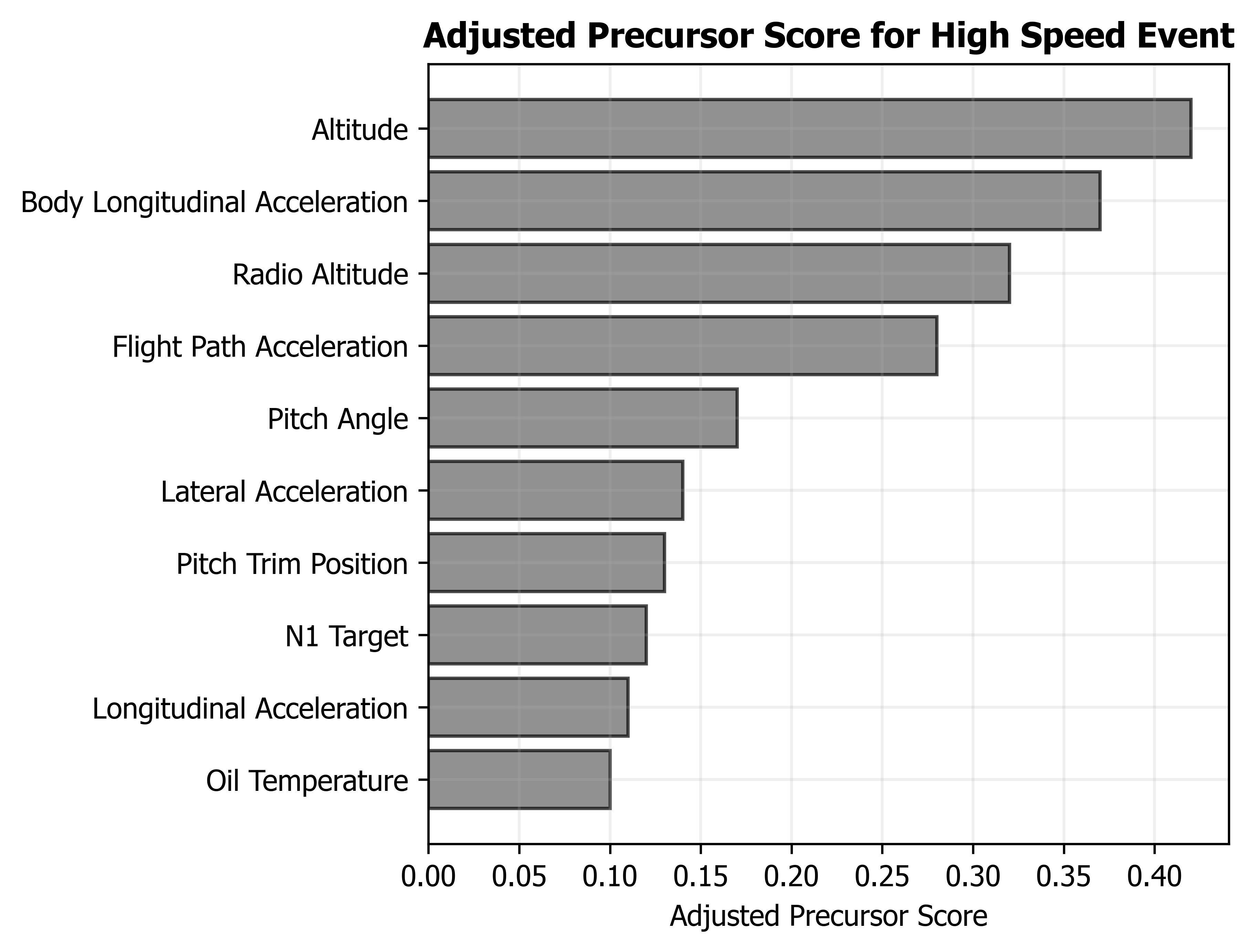}
	\caption{Precursor Ranking for a High Speed Event}
	\label{fig:adjusted_p_hs}
\end{figure}

When performing an inference using the trained model, the precursor scores can be extracted from the model's MHCNN outputs. Following previously highlighted steps in  \ref{subsub: adj precursor score}, the adjusted precursor score can be computed and used to identify the precursors for a flight. Fig. \ref{fig:adjusted_p_hs} shows the precursor ranking obtain for a flight that experienced a high speed event. Similar precursors were identified by ADOPT for this flight. The top 5 precursors were the altitude, the radio altitude, the flight path acceleration and the N1 target. Additionally, the the total pressure was also identified by ADOPT as a top precursor. Once the precursors are discovered, the aircraft's parameters can then be plotted to assess the them.

\begin{figure}[ht]
    \centering
	\includegraphics[width=.9\textwidth]{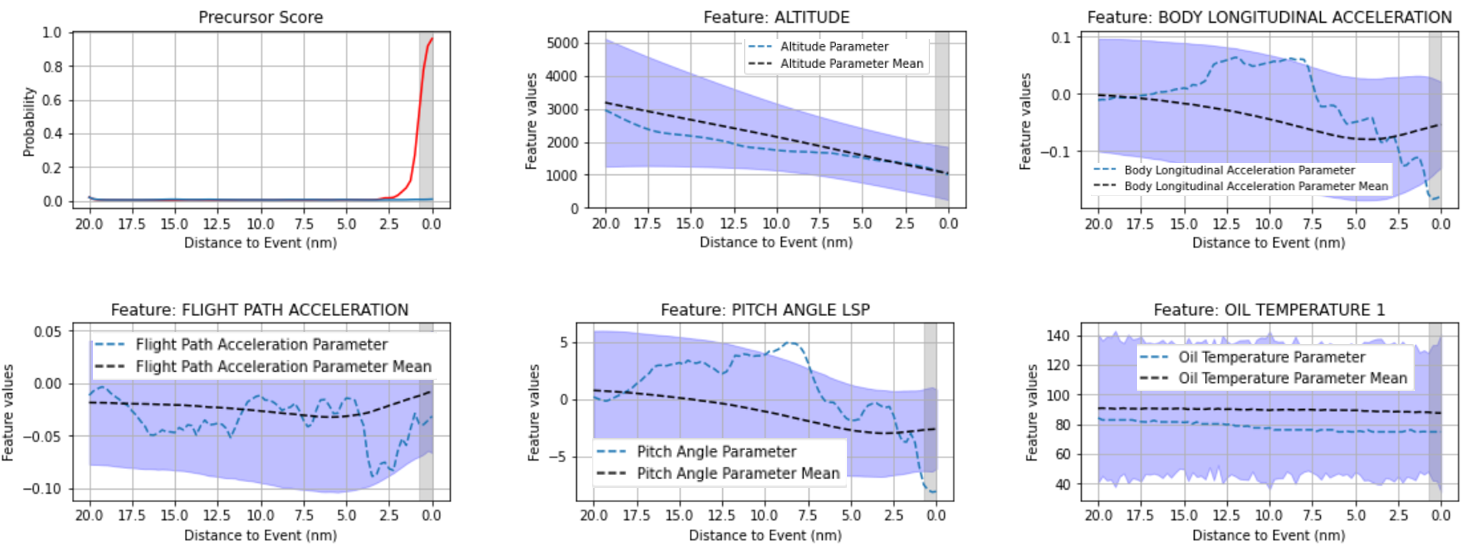}
	\caption{Precursor Score and Aircraft's Parameters during a High Speed Event}
	\label{fig:precursor_fl_data_hs}
\end{figure}

 In particular, the time series is of interest when the precursor score over time is greater than the $0.5$ threshold, which occurs at around 1.5 nautical miles away from a 1,000 ft above touch-down on fig. \ref{fig:precursor_fl_data_hs} (grey shaded area). On the figure, the red and blue curves in the first cell represent the precursor at different time steps obtained from the high speed and high path angle classifiers, respectively. Since the flying conditions do not correspond to a high path angle event, the classifier for the event cannot find precursors, while the classifier for the high speed event detects abnormal pattern leading to the identification of precursors. Note that if only the high path angle binary classifier was used, the model would have confidently classified the flight as normal. In the remaining cells, the dotted blue line represents the feature value. For instance this flight reached a pitch angle of 5 degrees 7.5 nautical miles away from a 1,000 ft. The dotted black line represents the mean nominal values for each feature at every time steps and the shaded purple region is defined as $\pm 2$ standard deviations away from the mean. From the figure and knowledge of the precursors it can be inferred that a potential cause to the excessive speed experienced by the flight at a 1,000 ft is correlated to the combination of negative accelerations (body longitudinal acceleration and flight path acceleration) and low pitch angle in the last 2.5 nautical miles. The oil temperature of the flight is also plotted to show that even though it has a lower values than the mean of nominal flights, the relatively low precursor score obtained is likely due to the fact that the temperature is relatively constant and is not highly correlated to the event.  

\subsubsection{High Path Angle Event Flight Analysis}\label{subsub:hp}
\begin{figure}[ht]
    \centering
	\includegraphics[width=.65\textwidth]{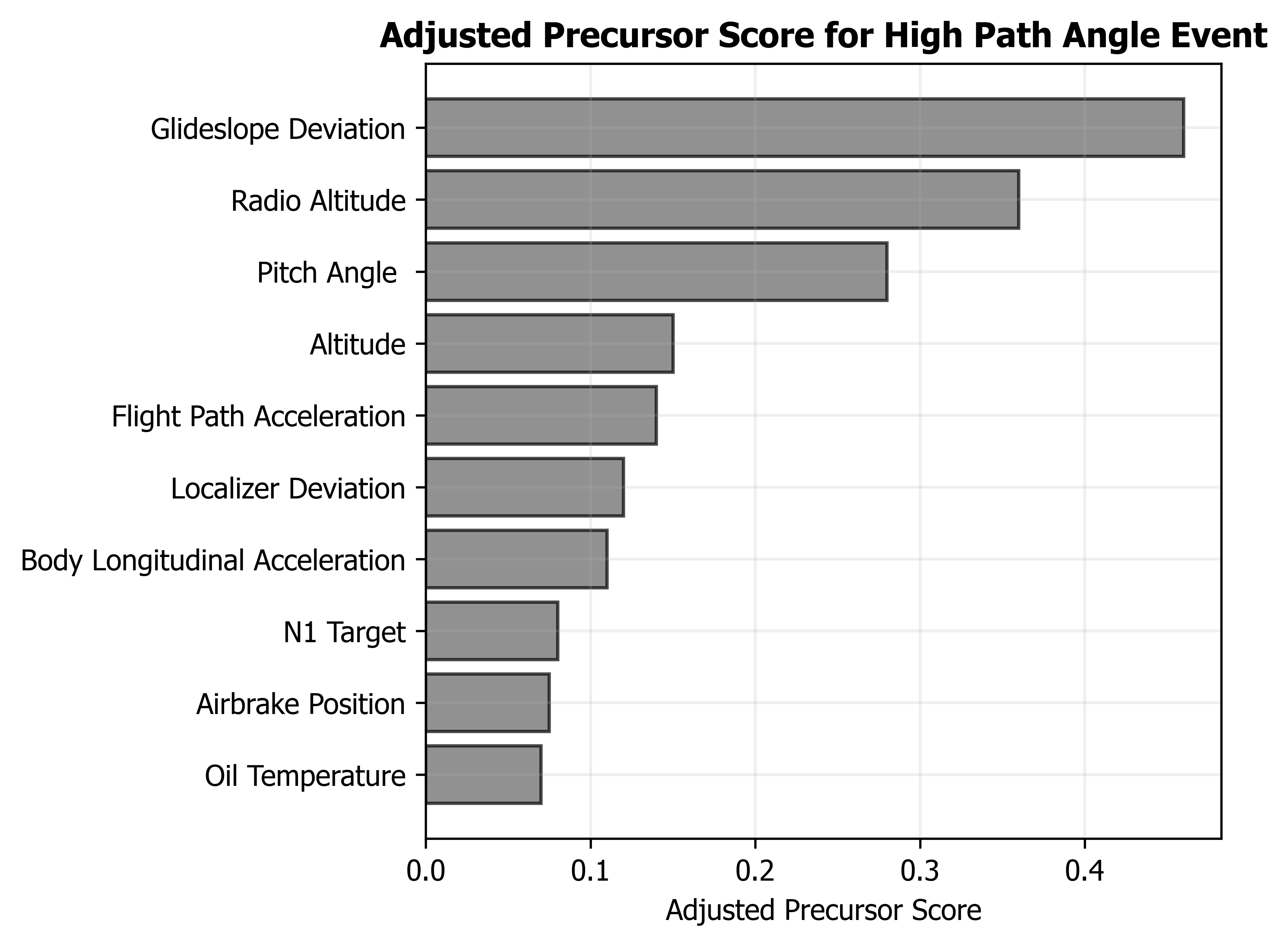}
	\caption{Precursor Ranking for a High Path Angle Event}
	\label{fig:adjusted_p_hp}
\end{figure}

Similar to the high speed event case, precursor to the high path angle event can be probed from the model. Again, the identified precursors provide a list of aircraft's parameters that can be assessed through visualization. For the selected flight the presence of a dominating precursor is observed. Indeed, there is a larger difference between the top two precursor than there was for the high speed event, as seen on fig. \ref{fig:adjusted_p_hp}. This larger difference suggest a highly abnormal glideslope deviation. Moreover, the glideslope was also identified by ADOPT as the top parameter. Other important parameters were the pitch angle, the radio altitude, the airbrake position, and the flight path acceleration. 

The abnormal behavior is confirmed by fig. \ref{fig:adjusted_p_hp} since the glideslope deviation is much greater than 2 times the standard deviation. The potential cause of the event is likely mainly related to the high deviation in the glideslope. However, other precursors such as the pitch angle and the flight path acceleration are also identified and observed to have abnormal patterns. It is important to note that these two parameters can also be precursors to the high speed event as previously observed. This flight in particular was not classified as a high speed event but some of the parameters behaved similarly to how they would behave during such event. This led the high speed event classifier to increase the precursor score over time probability to be closer to 1. Ultimately, the multi-class classifier correctly labeled this flight as a high path angle event because the high path angle classifier had a stronger confidence. 
\begin{figure}[ht]
    \centering
	\includegraphics[width=.9\textwidth]{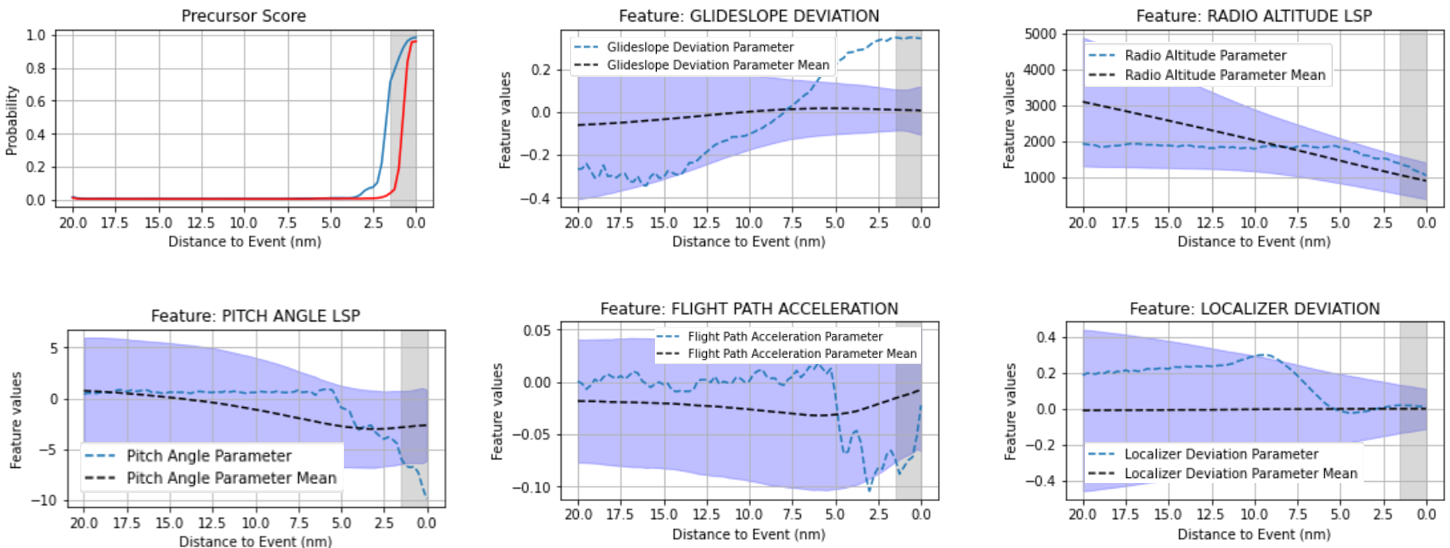}
	\caption{Precursor Score and Aircraft's Parameters during a High Path Angle Event}
	\label{fig:precursor_fl_data_hp}
\end{figure}

\subsection{Discussion of Results}
The MHCNN-RNN architecture yields satisfying results. The scores obtained from the conventional classification metrics show the model's ability to extract information from the data in order to make accurate predictions. In fact, the model accurately forecast the events since it predicts them before they actually occur, though variability in the prediction is observed. The model also identifies precursors, and it was observed that different precursors are discovered for different events. which is expected. The high speed top precursors relate more to poor energy management while the precursors for the high path angle include trajectory related parameters, as seen in table \ref{tab:precursor_score}. The identified precursors are partially validated given their resemblances with ADOPT identified precursors, which is characterized by the low DFA values. Additional support towards the validation of the discovered precursors is obtained through visualization and observing that the precursors exhibited abnormal behaviors, outside of nominal operations. 

\section{Conclusion}
The work presented in this paper tackles the precursor mining task. A public flight data set was leveraged to create FOQA-like labels. After performing the required preprocessing steps, a novel architecture for the task was developed to take advantage of the Multi-Instance Learning framework, the feature extraction capabilities of Convolutional Neural Networks and the temporal pattern recognition capabilities of Recurrent Neural Networks. On one hand binary classifiers were trained to predict both high speed and high path angle events, and on the other the MHCNN-RNN architecture was modified to handle multiple outputs. In both cases, a grid-search was implemented to determine the best parameters for the neural networks, and thus the best model for the prediction the safety events. For the multiple binary classifier case, the best two models were then combined to form an unique multi-class classifier. For both multi-class extensions, the final models were evaluated on a test set and high scores were observed for classification metrics such as F1 score, precision, and recall, in particular when combining binary classifiers. Furthermore, the binary models were then used to identify precursors and provide the average precursor score across all flights that experienced each of the two events. Finally, visualizations were used to observe the behaviors of the identified precursors, which exhibited patterns different from normal flight operations. Future work will include enhancing the interpretability of the precursor score tensor. While empirically the parameters that deviate from $0.5$ are understood to be correlated to the event of interest, the meaning of the direction of the deviation (greater than or lower than $0.5$) needs to be investigated. Further improvements to the model can be made towards increasing the prediction window, and allowing the classification of unknown precursors instead of defaulting unknown behaviors to the nominal class. Additionally, the lower scores for the modified MHCNN-RNN could be due to the class imbalance, especially the lower number of high path angle events. Future work will also explore methodologies to handle class imbalance, and extend the learning task to a multi-label problem.
% emperical results, need more on interpretation of the precursor score, improvement of performance for unseen event, end-to-end application for flight analysis , more events investigation, investigate other multi-class approach

\section*{Acknowledgments}
The authors would like to thank Dr. Nikunj Oza, Dr. Milad Memarzadeh, and Dr. Hamed Valizadegan for the feedback and insights they provided.
% TP Comment: Marc, can you explain the last sentence a little better? Does it mean that the bag-level label is applied to all instances within the bag?

\bibliography{sample}

\end{document}